\documentclass{ifcolog-c}
\usepackage[utf8]{inputenc}
\usepackage{graphicx}
\usepackage{url}

\newtheorem{example}{Example}

\begin{document}

\title{Evaluation of Automatically Constructed Word Meaning Explanations}

\titlethanks{This work has been partly supported by the Ministry of
    Education of CR within the LINDAT-CLARIAH-CZ project LM2018101.}
\titlerunning{Evaluation of Automatic Word Meaning Explanations}
\addauthor[\{413827,pary,hales\}@mail.muni.cz]{
  Marie~Stará,
  Pavel~Rychlý and
  Aleš~Horák%
}{%
  Faculty of Informatics
  Masaryk University\\
  Botanická 68a,
  Brno 602\,00\\
  Czech Republic%
}
\authorrunning{M.~Stará, P.~Rychlý, and A.~Horák}
\nopagenumber 
%
\maketitle

\begin{abstract}
Preparing exact and comprehensive word meaning explanations is one of the key steps in the process of monolingual dictionary writing. In standard methodology, the explanations need an expert lexicographer who spends a substantial amount of time checking the consistency between the descriptive text and corpus evidence.

In the following text, we present a new tool that derives explanations automatically based on collective information from very large corpora, particularly on word sketches. We also propose a quantitative evaluation of the constructed explanations, concentrating on explanations of nouns. The methodology is to a certain extent language independent; however, the presented verification is limited to Czech and English. 

We show that the presented approach allows to create explanations that contain data useful for understanding the word meaning in approximately 90\% of cases. However, in many cases, the result requires post-editing to remove redundant information.
\def\and{\unskip,\ }
\keywords{explanations  \and word sketches \and explanation construction}
\end{abstract}

\section{Introduction}

When an expert lexicographer constructs a (monolingual) dictionary, one of the most challenging and time-consuming tasks is to create concise and comprehensive word meaning explanations, also referred to as (dictionary) definitions~\cite{gilliver2016making,gray1986creating,mckeown1993creating}. The standard approaches concentrate on selecting the shared vocabulary of terms used to describe the word and organize them in the order of the main word category followed by listing the characteristics which are specific to the word~\cite{svensen2009handbook}.

In this paper, we describe a new attempt to develop dictionary word explanations for Czech and English automatically, using statistical information aggregated from large text corpora. We work with the hypothesis that a meaning of a word can be deduced from its context~\cite{church2011pendulum}. Therefore, it is possible to abstract common collocations of a word and use them to explain the word meaning. Such an explanation helps the reader to understand a meaning of a new word unfamiliar beforehand.


In the following section, we discuss the related work and the uniqueness of the presented approach. In sections three and four, we discuss the method and evaluate the results. Section five concludes the text.

\section{Related Work}

The attempts to actually \emph{create} explanations automatically have been rather scarce.  Labropoulou et al.~\cite{CLgen} generated dictionary definitions from a computational lexicon, i.e.\ a lexicon of formalized and explicitly encoded semantic information about words. The results were comprehensible and the generated definitions were consistent, the downside being the need for the ontological background in the form of the computational lexicon. That is why the authors were focusing only on selected concrete entities; our aim is to cover a broader part of the vocabulary. 

The automated definition construction process has been mostly solved
by text mining approaches. There were attempts to \emph{find},
\emph{mine}, or \emph{extract} definitions. Early approaches, such
as~\cite{zdipl3}, used rule-based or pattern-based approaches to
identify text passages containing the sought term and its explanation.
Such an approach is usually limited to a selected domain and sources of
texts to allow for acceptable precision and recall. The pattern-based
approach was later adjusted for mining from very large
corpora~\cite{zdipl4} which offered improved precision of
\mbox{73--74\%} with Wikipedia corpora and 31--57\% with large web
corpora. Borg et al.~\cite{zdipl1} employed genetic programming
techniques to generate the best definition templates and to learn to
rank these templates by importance. The templates were then used to
identify definitions in non-technical texts with high precision (up to
100\%) but with about 50\% recall. Later works~\cite{zdipl5,zdipl2} solved the definition text search by
annotating a corpus of definitions (from Wikipedia or from scientific
papers) and then training a sequence labeling classifier to mark words
as \emph{term}, \emph{definition} or \emph{other}. This technique
improved the F-score with Wikipedia benchmark corpus to 85\%.

All these techniques concentrate on extracting (parts of) the
definitions from existing human-made texts.  Such an approach is
useful for summarization of technical terminology and educative texts,
but not for general notion explanations. Another possible problem with
the extractive approaches lies in the authorship laws and their
possible breach. 

In the following text, we concentrate on extending our previous work
published by Stará and Kovář~\cite{staraslanen,staraslancs,svojtou}.
The method and the involved tools are described in detail here and
a quantitative evaluation of explanations is offered.

\section{Method}

The presented explanation creation method has been evaluated with nouns, adjectives and verbs in the Czech and English languages using Word Sketches~\cite{ske10,ske04,skegram} compiled with specific sketch grammars and the \emph{csTenTen12}\footnote{\url{https://www.sketchengine.eu/cstenten-czech-corpus/}} and \emph{enTenTen13}\footnote{\url{https://www.sketchengine.eu/ententen-english-corpus/}} corpora provided by the Sketch Engine corpus management system.$\!$\footnote{Sketch Engine is a tool analysing text corpora to identify instantly what is typical in language and what is rare, unusual or emerging usage. See \url{https://www.sketchengine.eu/} for details.} 

\begin{figure}[t]
    \begin{center}
        \includegraphics[width=.8\textwidth]{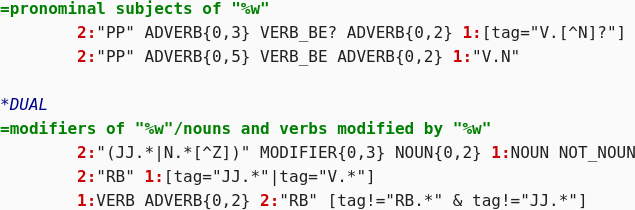}
    \end{center}
    \caption{An example of sketch grammar relations of
    \emph{pronominal subjects} and \emph{modifiers of a word}}
    \label{fig_gram}
\end{figure}

\begin{figure}[t]
\includegraphics[width=1\textwidth]{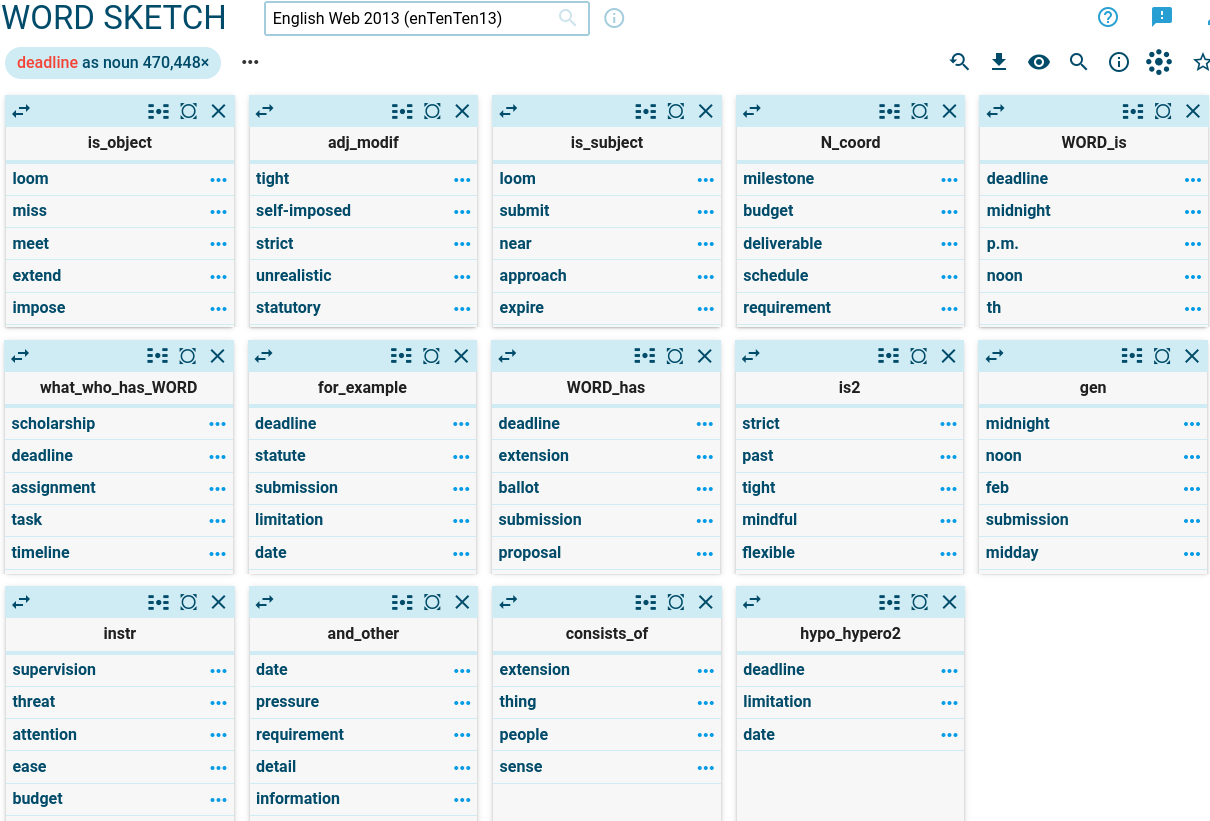}
\caption{Word sketches for the word \emph{deadline}} \label{fig1}
\end{figure}

A sketch grammar is a set of syntactic queries written in the corpus query language (CQL~\cite{cql}) to identify inter-word relations based on their position, distance, part-of-speech tag and word form (see Figure~\ref{fig_gram} for an example). The grammar, i.e.\ the set of word relation rules, can be applied to large text corpus to create the word sketches. The word sketches show statistically meaningful collocations of a given word organized by the relation rules; an example is shown in Figure~\ref{fig1}.

The definition grammars are partially based on existing grammars for the above-mentioned corpora. Specific modifications are focused mainly on identifying hypernyms/hyponyms and meronyms/holonyms for nouns, opposites and specific noun collocations for adjectives, and prepositional phrases for verbs. 
Apart from word sketches, we also employ the thesaurus tool~\cite{ske10,ske04} to obtain synonyms.

These tools provide the grounding of the information that should be contained in the explanation. Following the explanation schemata by renowned dictionary creation guidelines~\cite{ogpl,landau,ml,plr}, we have compiled a list of definition/ex\-pla\-na\-tion types. In the overview below, we show the links between the types and the sketch grammar relations.

The standard way of how to explain a word meaning consists of two parts: the \emph{genus proximum} and \emph{differentia specifica}. In other words, using a hypernym as the core determination and a set of features that distinguish the word from other related words. As long as the headword is a (concrete) noun, there is quite a clear path to a hypernym. Identification of the distinguishing features is, however, not so straightforward. They can have the form of a \emph{verb} (``dog: an animal that barks''), a \emph{noun} describing what the headword has got (``snake: an animal with forked tongue'') or what the headword lacks (``snake: an animal with no legs''). In this regard, meronyms and holonyms are a specific case (``cutlery: spoon, fork, knife'').

Another approach to explaining lies in using the \emph{ostensive definition} or explaining by pointing. Pointing is quite helpful when explaining adjectives (``blue as the sky''). For adjective definitions, opposites can also bring clarification (``dead: not alive'').

Using synonyms for explaining is not much encouraged; however, we find synonymy (or, in case of verbs, troponymy) generally useful. Here, we should remark that we consider a synonym quite loosely, taking into account any words that have a similar meaning.

To describe the meaning of \emph{verbs}, we mostly make use of valency, focusing especially on objects. 
We use valency with other parts-of-speech as well, mainly to find nouns and adjective modifiers.

\begin{figure}[t]
\noindent
    \begin{minipage}{\textwidth}
        \begin{quote}
            \emph{\textbf{bone}}:
            \begin{enumerate}
                \item similar meaning as a/an \textbf{bone} can have (a/an) \emph{tooth}, \emph{joint}, \emph{muscle}, \emph{tissue}, \emph{fracture}, \emph{calcium}, \emph{osteoporosis}, \emph{skull}, \emph{spine}, \emph{injury}, \emph{remain}
                \item \textbf{bone} can be \emph{bare}, \emph{pubic}, \emph{brittle} 
                \item for example (a/an) \emph{femur}, \emph{vertebra}
                \item \textbf{bone} can have/contain (a/an) \emph{marrow}, \emph{skull}, \emph{joint}, \emph{tooth}
                \item (a/an) \emph{tissue}, \emph{osteoporosis} can have/contain (a/an) \textbf{bone}
                \item \textbf{bone} can \emph{fragment}, \emph{heal}, \emph{fracture}
                \item sth/sb can \emph{break}, \emph{strengthen}, \emph{fracture} a/an \textbf{bone}
                \item \textbf{bone} of (a/an) \emph{contention}, \emph{skull}, \emph{spine}
                \item \textbf{bone} with (a/an) \emph{flesh}, \emph{marrow},
                    \emph{meat}
            \end{enumerate}
        \end{quote}
    \end{minipage}
    \caption{An example automatic explanation of the word \emph{bone}}
    \label{fig_bone}
\end{figure}

To create the explanation as such, we use an automated script to combine all results together, with the main source being the word sketches and the thesaurus providing extra synonyms. The word sketches are sorted by their frequency score limited to the first three results. Some of the sketch relations are merged to one line in the explanation while removing duplicates. An example explanation for the word \emph{bone} is presented in Figure~\ref{fig_bone}. The lines are enumerated to make referencing easier.

The first line joins results from several relations~--~\emph{cooperation}, \emph{hypernymy}, and results from the \emph{thesaurus}. The second line introduces significant \emph{adjective modifiers}. The third line shows \emph{examples} of the headword. The fourth and fifth lines present the \emph{meronyms} and \emph{holonyms}, respectively.
The sixth and seventh lines list \emph{verbs} that typically have the headword as a \emph{subject} and \emph{object}. The eighth and ninth line post \emph{nouns} connected with the headword by \emph{genitive} and \emph{instrumental} case.

\section{Evaluation}

For both Czech and English, we manually evaluated a test set consisting of 71~nouns, 33~adjectives and 40~verbs.$\!$\footnote{42 for Czech due to aspect variants.}

\begin{table}[t]
\centering
    \caption{Indicators in Czech and English Explanations}
    \label{cs_en_markers}
\tiny
\begin{minipage}{.47\textwidth}
    \centering
    Indicators in Czech Explanations

    \begin{tabular}{lrrr}
        \hline
                    & N~~~    & J~~~    & V~~~     \\ \hline
        synonym     & 92.96\% & 78.79\% & 90.48\%  \\
        J modifier  & 95.77\% & -       & -        \\
        subject     & 97.18\% & -       & 90.48\%  \\
        object      & 94.37\% & -       & 100.00\% \\
        hypernym    & 67.61\% & -       & -        \\
        hyponym     & 29.58\% & -       & -        \\
        meronym     & 54.93\% & -       & -        \\
        holonym     & 46.48\% & -       & -        \\
        A modifier  & -       & -       & 90.48\%  \\
        (such) as   & -       & 36.36   & -        \\
        troponym    & -       & -       & 45.24\%  \\
        opposite    & -       & 63.64\% & -        \\
        PP          & -       & -       & 97.62\% \\
        infrequent  & 54.93\% & 57.58\% & 45.24\% \\
        data issues & 81.69\% & 30.30\% & 95.86\% \\\hline
    \end{tabular}
\end{minipage}\hspace{6mm}%
\begin{minipage}{.47\textwidth}
    \centering
    Indicators in English Explanations

    \begin{tabular}{lrrr}
    \hline
                    & N~~~    & J~~~    & V~~~    \\\hline
        synonym     & 92.96\% & 72.73\% & 85.00\% \\
        Jmodifier   & 87.32\% & -       & -       \\
        subject     & 69.01\% & -       & 85.00\% \\
        object      & 80.28\% & -       & 87.50\% \\
        hypernym    & 54.93\% & -       & -       \\
        hyponym     & 21.13\% & -       & -       \\
        meronym     & 56.34\% & -       & -       \\
        holonym     & 56.34\% & -       & -       \\
        Amodifier   & -       & 75.76\% & 9.00\% \\
        as          & -       & 42.42\% & -       \\
        troponym    & -       & 0.00\%  & 52.50\% \\
        opposite    & -       & 45.45\% & -       \\
        PP          & -       & -       & - \\
        infrequent  & 61.97\% & 18.18\% & 35.00\% \\
        data issues & 12.68\% & 36.36\% & 32.50\% \\\hline
    \end{tabular}
\end{minipage}

\medskip
N: noun, J: adjective, V: verb, A: adverb, PP: prepositional phrase
\end{table}

The evaluation proceeded in a quantitative way measuring the occurrence numbers of identified features. The resulting aggregated score should correspond to the decision about the usefulness of the explanation. The indicators of a (presumably) good explanation are the presence of (useful):
\begin{itemize}
    \item synonyms: all parts-of-speech
    \item adjective modifiers: nouns
    \item adverbial modifiers: verbs, (adjectives)
    \item noun collocation: adjectives (\textit{(such) as}), verbs (\textit{is subject/object of})
    \item verb collocation: nouns (\textit{subject/object})
    \item opposites: adjectives
    \item hypernyms/hyponyms: nouns
    \item meronyms/holonyms (part of): nouns
    \item troponyms: verbs
    \item prepositional phrases: verbs
\end{itemize}
On the other hand, certain features may also serve as negative indicators.
Markers of possible problems with the explanation are the presence of:
\begin{itemize}
    \item infrequent expressions
    \item errors caused by the corpus data (wrong lemma/tag;
        interchanging objects and subjects, meronyms and holonyms,
        etc.)
\end{itemize}
Table~\ref{cs_en_markers} lists the ratios of explanations that contain the said (positive or negative) indicator. The fact that the indicator is not present does not necessarily mean that the explanation is bad or insufficient: different words require different indicators, as discussed in Section~\ref{Neval}.

Table~\ref{overviev} shows the total number of explanations that are sufficient as-is, denoted as \emph{good} explanations. The explanations that contain \emph{some} of the important data but are either incomplete (lack some of the necessary information) or contain too much junk data, or the data are misleading (e.g.\ opposites are presented as synonyms) are counted as \emph{post-edit}; the last group of explanations, \emph{bad}, consists of those that are completely insufficient or contain so many issues they would require rewriting, not just post-editing.

As the results were evaluated manually, we necessarily used our subjective view towards the evaluation based on our experience as a language and dictionary users. Nevertheless, we tried to minimize the bias by following the above-mentioned indicators.
We plan to engage more evaluators in the future to offer a broad objective assessment of the quality and intelligibility of the explanations.

\begin{table}[t]
\caption{Overall quality of Explanations}
\label{overviev}
\centering
\begin{tabular}{lrr@{\quad}r}
\hline
Czech & good & ~~post-edit & bad \\\hline
N  &   39.44\%   &    52.11\%          &    8.45\% \\
J  &    42.42\%  &    27.27\%          &    30.30\% \\
V  &    16.67\%  &    78.57\%          &    4.76\% \\\hline
\end{tabular}\hspace{8mm}%
\begin{tabular}{lrr@{\quad}r}
\hline
English & good & ~~post-edit & bad \\\hline 
N  &   39.44\%   &    49.30\%          &    11.27\% \\
J  &    21.21\%  &    60.61\%          &    18.18\% \\
V  &    23.81\%  &    69.05\%          &    2.38\% \\\hline
\end{tabular}

\medskip
N: noun, J: adjective, V: verb
\end{table}

\subsection{Nouns}\label{Neval}

In this section, we present a more detailed evaluation of noun explanations, offering a comparison with existing dictionary definitions in the Macmillan Dic\-tio\-nary.$\!$\footnote{\url{https://www.macmillandictionary.com/}}

A universally acknowledged truth says that a noun explanation should contain its hypernym. Even though this is generally true, as e.g.\ in Example~\ref{ex1} below, there are counterexamples, such as Examples~\ref{ex2} and~\ref{ex3} where the hypernym is too general or replaced by a synonym, respectively. When evaluating the testing dataset with the established dictionary, we see that a hypernym is present in 70.42\% of the noun definitions, while 4.23\% headwords are not defined in the dictionary.
\begin{quote}

    \begin{example}\label{ex1}
    \textbf{\emph{deer}}: a large brown \textbf{animal} with long thin legs. The adult male \textbf{deer} is called a stag and may have antlers growing from its head. The female \textbf{deer} is called a doe and a young \textbf{deer} is called a fawn.\footnote{\url{https://www.macmillandictionary.com/dictionary/british/deer}}
    \end{example}

    \begin{example}\label{ex2}
    \emph{\textbf{teacher}}: someone whose job is to \textbf{teach}\footnote{\url{https://www.macmillandictionary.com/dictionary/british/teacher}}
    \end{example}

    \begin{example}\label{ex3}
    \emph{\textbf{stream}}: a small narrow \textbf{river}\footnote{\url{https://www.macmillandictionary.com/dictionary/british/stream_1}}
    \end{example}

\end{quote}
Examples~\ref{ex1}, \ref{ex2}, and~\ref{ex3} are all cases of a good definition, as they denote what does the headword mean. It is important to note that all these explanations use different strategies of what semantic relations to use. 

In the second example, a verb describing the prototypical activity of the headword is necessary, while in the third example, only a synonym with a few modifiers is sufficient. Such an approach is not always applicable, as can be seen in Example~\ref{ex4}. An explanation like this could be as well used to describe a \emph{shrew}, \emph{rat}, or \emph{opossum}, even a \emph{cat}. To avoid such confusion, we decided to prefer redundant data over data scarcity.
\begin{quote}

    \begin{example}\label{ex4}
    \emph{\textbf{mouse}}: a small furry \textbf{animal} with a long tail\footnote{\url{https://www.macmillandictionary.com/dictionary/british/mouse_1}}
    \end{example}


\end{quote}
Example~\ref{ex5} shows an automatically created explanation that can be compared with the human-made one. The explanation contains hypernyms (\emph{water, waterway}; possibly also \emph{source, body}); synonyms (\emph{river, tributary}; possibly also \emph{lake, pond, channel}; and verb collocates for \emph{\textbf{stream}} as a subject (\emph{flow, meander}). 
\begin{quote}

    \begin{example}\label{ex5}\hfuzz=25pt
    \emph{\textbf{stream}}:
    \begin{itemize}
    \item similar meaning as a/an \textbf{stream} can have (a/an) \emph{river}, \emph{lake}, \emph{pond}, \emph{flow}, \emph{channel}, \emph{tributary}, \emph{water}, \emph{source}, \emph{waterway}, \emph{body}
    \item \textbf{stream} can be \emph{steady}, \emph{endless},
        \emph{constant} 
    \item for example (a/an) \emph{river}, \emph{habitat}
    \item \textbf{stream} can have/contain (a/an) \emph{trout}, \emph{flow}, \emph{waterfall}
    \item (a/an) \emph{watershed}, \emph{valley} can have/contain (a/an) \textbf{stream}
    \item \textbf{stream} can \emph{flow}, \emph{meander}, \emph{replenish}
    \item sth/sb can \emph{cross}, \emph{never-end}, \emph{flow} a/an \textbf{stream}
    \item \textbf{stream} of (a/an) \emph{income}, \emph{consciousness}, \emph{revenue}
    \item \textbf{stream} with (a/an) \emph{waterfall}, \emph{trout}
    \end{itemize}
    \end{example}
\end{quote}

\noindent
The results for nouns are encouraging, as a significant number of the explanations helps to understand the word meaning without the need of \emph{excessive} post-editing. The other parts of speech seem to require a slightly different approach, mainly if we compare the explanations to existing dictionary definitions. As adjectives' main function is to modify, we need to change the paradigm and accept the fact that the results can be helpful even when they do not conform to the standard definitions. A similar problem occurs with verbs.

\section{Conclusions}

In this paper, we introduced a new tool for automatic construction of word meaning explanations for Czech and English, using large corpora, especially the word sketches technique. We have conceived a quantitative evaluation of explanations, focusing mainly on explanations of nouns.

The presented approach gathers enough data to construct explanations for 91.25\% and 88.73\% of nouns for Czech and English, respectively. As a majority of the results needs post-editing, the output in general is not yet in the state that could be presented to users as actual explanations. However, the status quo can be used as a basis for human-made explanations or definitions.

To further improve our work, the next tasks will be finding out which words need which specific approach, such as deciding which words do (not) require a hypernym, or for what words it is necessary to output a verb collocation. We believe such steps will further improve the results and reduce the need for post-editing.



\bibliographystyle{plain}
\bibliography{main}
\end{document}